\documentclass[11pt,letterpaper]{article}
\usepackage{emnlp2017}
\usepackage{times}
\usepackage{latexsym}
\usepackage{array}
\usepackage{graphicx}
\usepackage{multirow}
\usepackage{amsmath}
\usepackage{subfig}
\usepackage{mathtools}
\usepackage{dsfont}

\emnlpfinalcopy

\newcommand\MT{\texttt{MT}}
\newcommand\PE{\texttt{PE}}
\newcommand\OP{\texttt{OP}}
\newcommand\SRC{\texttt{SRC}}
\newcolumntype{C}[1]{>{\centering\let\newline\\\arraybackslash\hspace{0pt}}m{#1}}
\newcolumntype{F}{>{\centering\let\newline\\\arraybackslash\hspace{0pt}}m{1.8cm}}

\title{LIG-CRIStAL System for the WMT17 Automatic Post-Editing Task}

\author{Alexandre B\'erard \hspace{0.3cm} Olivier Pietquin\thanks{*now with DeepMind, London, UK} \\
        Univ. Lille, CNRS, Centrale Lille, Inria, UMR 9189 CRIStAL\\
        {\tt \normalsize alexandre.berard@ed.univ-lille1.fr} \\ {\tt \normalsize olivier.pietquin@univ-lille1.fr}
        \AND Laurent Besacier \\
        LIG, Univ. Grenoble Alpes, CNRS \\
        {\tt \normalsize laurent.besacier@univ-grenoble-alpes.fr}}

\date{}

\begin{document}

\maketitle

\begin{abstract}
    This paper presents the LIG-CRIStAL submission to the shared Automatic Post-Editing task of WMT 2017.
    We propose two neural post-editing models: a mono-source model with a task-specific attention mechanism, which performs particularly well in a low-resource 
    scenario; and a chained architecture which makes use of the source sentence to provide extra context. This latter architecture manages
    to slightly improve our results when more training data is available. We present and discuss our results on two datasets (\emph{en-de} and \emph{de-en})
    that are made available for the task.
\end{abstract} 

\section{Introduction}
\label{sec:intro}

It has become quite common for human translators to use machine translation (MT) as a first step, and then to manually post-edit the translation hypothesis. This can result in a significant gain of time, compared to translating from scratch \cite{green2013}.
Such translation workflows can result in the production of new training data, that may be re-injected into the system in order to improve it. Common ways to do so are retraining, incremental training, translation memories, or automatic post-editing \cite{chatterjee2015}.

In Automatic Post-Editing (APE), the MT system is usually considered as a blackbox: a separate APE system takes as input the outputs of this MT system, and tries to improve them.
Statistical Post-Editing (SPE) was first proposed by \newcite{simard2007}. It consists in training a Statistical Machine Translation (SMT) system \cite{moses}, to translate from translation hypotheses to a human post-edited version of those. \newcite{bechara2011} then proposed a way to integrate both the translation hypothesis and the original (source language) sentence. More recent contributions in the same vein are \cite{chatterjee2016,pal2016}.

When too little training data is available, one may resort to using synthetic corpora: with simulated PE \cite{potet2012}, or round-trip translation \cite{amu}.

Recently, with the success of Neural Machine Translation (NMT) models \cite{sutskever2014,bahdanau2015}, new kinds of APE methods have been proposed that use encoder-decoder approaches \cite{amu,amu2017,cuni,pal2017,hokamp}, in which a Recurrent Neural Network (RNN) encodes the source sequence into a fixed size representation (encoder), and another RNN uses this representation to output a new sequence.
These encoder-decoder models are generally enhanced with an attention mechanism, which learns to look at the entire sequence of encoder states \cite{bahdanau2015,luong2016}.

We present novel neural architectures for automatic post-editing. Our models learn to generate sequences of edit operations, and use a task-specific attention mechanism which gives information about the word being post-edited.

\subsection{Predicting Edit Operations}

We think that post-editing should be closer to spelling correction than machine translation. Our work is based on \newcite{cuni}, who train a model to predict edit operations instead of words. We predict 4 types of operations: \texttt{KEEP}, \texttt{DEL}, \texttt{INS(word)}, and \texttt{EOS} (the end of sentence marker). This results in a vocabulary with three symbols plus as many symbols as there are possible insertions.

A benefit of this approach is that, even with little training data, it is very straightforward to learn to output
the translation hypothesis as is (MT baseline). We want to avoid a scenario where the APE system is weaker than the original MT system and only degrades its
output.
However, this approach also has shortcomings, that we shall see in the remainder of this work.

\paragraph{Example}

If the \MT{} sequence is \texttt{"The cats is grey"}, and the output sequence of edit ops is \texttt{"KEEP DEL INS(cat) KEEP KEEP INS(.)"},
this corresponds to doing the following sequence of operations:
keep \texttt{"The"}, delete \texttt{"cats"}, insert \texttt{"cat"}, keep \texttt{"is"}, keep \texttt{"grey"}, insert \texttt{"."}
The final result is the post-edited sequence \texttt{"The cat is grey ."}

We preprocess the data to extract such edit sequences by following the shortest edit path (similar to a Levenshtein distance, without substitutions, or with a substitution cost of $+\infty$).

\subsection{Forced Attention}

State-of-the-art NMT systems \cite{bahdanau2015} learn a global attention model, which helps the decoder look at the relevant part of the input sequence each time it generates a new word. It is defined as follows:

\vspace{-1cm}
\begin{align}
    \label{eq:global}
    attn_{global}(\mathrm{h}, s_t) = \sum_{i=1}^{A}a_i^t h_i \\
    a_i^t = softmax(e_i^t) \\
    e_i^t = v^Ttanh(W_1h_i+W_2s_t + b_2)
\end{align}

\noindent where $s_t$ is the current state of the decoder, $h_i$ is the i\textsuperscript{th} state of the encoder (corresponding to the i\textsuperscript{th} input word). $A$ is the length of the input sequence. $W_1$, $W_2$ and $b_2$ are learned parameters of the model. This attention vector is used to generate the next output symbol $w_t$ and to compute the next state of the decoder $s_{t+1}$.

However, we don't predict words, but edit operations, which means that we can do stronger assumptions as to how the output symbols align with the input.
Instead of a soft attention mechanism, which can look at the entire input and uses the current decoder state $s_t$ to compute soft weights $a_i$;
we use a hard attention mechanism which directly aligns $t$ with $i$. The attention vector is then $attn_{forced}(h,s_t)=h_i$.

The $t\rightarrow i$ alignment is pretty straightforward: $i$ is the number of \texttt{KEEP} and \texttt{DEL} symbols in the decoder's past output $(w_1,\dots,w_{t-1})$ plus one.

Following the example presented earlier, if the decoder's past output is \texttt{"KEEP DEL INS(cat)"}, the next token to generate is naturally aligned with the third input word ($i=3$), i.e., we've kept \texttt{"The"} and replaced \texttt{"cats"} with \texttt{"cat"}. Now, we want to decide whether we keep the third input word \texttt{"is"}, delete it, or insert a new word before it.

If the output sequence is too short, i.e., the end of sentence marker \texttt{EOS} is generated before the pointer $i$ reaches the end of the input sequence, we automatically pad with \texttt{KEEP} tokens. 
This means that to delete a word, there must always be a corresponding \texttt{DEL} symbol. This ensures that, even when unsure about the length of the output sequence, the decoder remains conservative with respect to the sequence to post-edit.

\subsection{Chaining Encoders}

The model we proposed does not make any use of the source side \SRC{}. Making use of this information is not very straightforward in our framework.
Indeed, we may consider using a multi-encoder architecture \cite{zoph2016,amu2017}, but it does not make much sense to align an edit operation with the source sequence, and such a model struggles to learn a
meaningful alignment.

\begin{figure}
    \hspace{-.2cm}
    \includegraphics[width=0.5\textwidth]{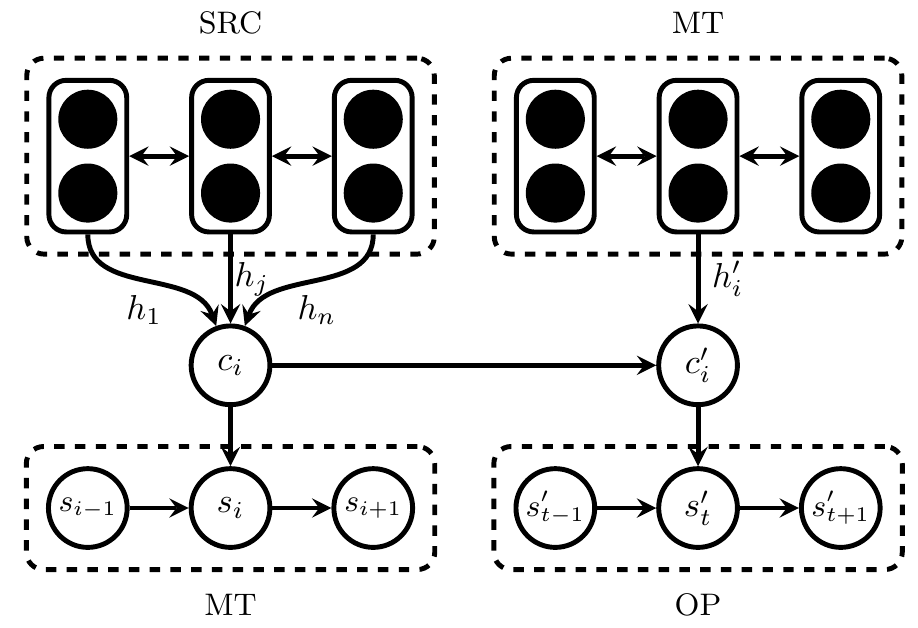}
    \caption{There are two bidirectional encoders that read the \SRC{} and \MT{} sequences. We maximize two training objectives: a translation objective ($\SRC\to\MT$) and a post-editing objective ($\MT\to\OP$). The \OP{} decoder does a forced alignment with the \MT{} encoder ($t\to i$), and uses the corresponding global attention context $c_i$ over \SRC{}: $c'_i=tanh(H_1 c_i + H_2 h'_i + b')$.
    The \MT{} decoder and \MT{} encoder share the same embeddings.}
    \label{fig:chained}
\end{figure}

We propose a chained architecture, which combines two encoder-decoder models (see fig.~\ref{fig:chained}). A first model $\SRC\to\MT$, with a global attention mechanism, tries to mimic the translation process that produced \MT{} from \SRC{}. The attention vectors of this first model summarize the part of the \SRC{} sequence that led to the generation of each \MT{} token.
A second model $\MT\to\OP$ learns to post-edit and uses a forced attention over the \MT{} sequence, as well as the attention vectors over \SRC{} computed by the first system.
Both models are trained jointly, by optimizing a sum of both losses.

\section{Experiments}

\begin{table}
    \centering
    \begin{tabular}{ | c | c | c | c | c | }
        \hline
        \multirow{2}{*}{Task} & \multirow{2}{*}{Train} & \multirow{2}{*}{Dev} & Test & \multirow{2}{*}{Extra} \\
                              &                        &                      & 2016 &                        \\
        \hline
        \multirow{2}{*}{\emph{en-de}} & 23k & \multirow{2}{*}{1000} & \multirow{2}{*}{2000} & 500k \\
                                      & (12k + 11k) &                       &                       & 4M   \\
        \hline
        \emph{de-en} & 24k & 1000 & none & none \\
        \hline
    \end{tabular}
    \caption{Size of each available corpus (number of \SRC{}, \MT{}, \PE{} sentence tuples).}
    \label{table:corpora}
\end{table}

This year's APE task consists in two sub-tasks: a task on English to German post-editing in the IT domain (\emph{en-de}), and a task on German to English post-editing in the medical domain (\emph{de-en}).
Table~\ref{table:corpora} gives the size of each of the corpora available.
The goal of both tasks is to minimize the HTER \cite{hter} between our automatic post-editing output, and the human post-editing output.

The \emph{en-de} \emph{23k} training set is a concatenation of last year's \emph{12k} dataset, and a newly released \emph{11k} dataset. A synthetic corpus was built and used by the winner of last year's edition \cite{amu}, and is available this year as additional data (\emph{500k} and \emph{4M} corpora). 

For the \emph{en-de} task, we limit our use of external data to the 500k corpus. For the \emph{de-en} task, we built our own synthetic corpus, using a technique similar to \cite{amu}.

\subsection{Synthetic Data}

\paragraph{Desiderata}
We used similar data selection techniques as \newcite{amu}, applied to the \emph{de-en} task.
However, we are very reticent about using as much parallel data as the authors did. We think that access to such amounts of
parallel data is rarely possible, and the round-trip translation method they used too cumbersome and unrealistic.
To show a fair comparison, this paper should show APE scores when translating from scratch with an MT system trained with all this parallel data.

To mitigate this, we decided to limit our use of external data to monolingual English (\emph{commoncrawl}). So, the only parallel data we 
use is the \emph{de-en} APE corpus.

\paragraph{PE side}

Similarly to \newcite{amu} we first performed a coarse filtering of well-formed sentences of commoncrawl. After this filtering step, we obtained about 500M lines.
Then, we estimated a trigram language model on the \PE{} side of the APE corpus, and sorted the 500M lines according to their log-score divided by sentence length.
We then kept the first 10M lines. This results in sentences that are mostly in the medical domain.

\paragraph{MT and SRC sides}
Using this English corpus, and assuming its relative closeness to the \PE{} side of the APE corpus, we now need to generate \SRC{} and \MT{} sequences.
This is where our approach differs from the original paper.

Instead of training two SMT systems $\PE\to\SRC$ and $\SRC\to\MT$ on huge amounts of parallel data, and doing a round-trip translation of the monolingual data,
we train two small $\PE\to\SRC$ and $\PE\to\MT$ Neural Machine Translation systems on the APE data only.

An obvious advantage of this method is that we do not need external parallel data. The NMT systems are also fairly quick to train, and evaluation is very fast.
Translating 10M lines with SMT can take a very long time, while NMT can translate dozens of sentences at once on a GPU.

However, there are strong disadvantages:
for one, our \SRC{} and \MT{} sequences have a much poorer vocabulary as those obtained with round-trip translation (because we only get words that
belong to the APE corpus). Yet, we hope that the richer target (\PE{}) may help our models learn a better language model.

\paragraph{TER filtering} Similarly to \newcite{amu}, we also filter the triples to be close to the real \PE{} distribution in terms of TER statistics. We build a corpus of the \emph{500k} closest tuples.
For each tuple in the real PE corpus, we select a random subset of \emph{1000} tuples from the synthetic corpus and pick the tuple whose euclidean distance with the real PE tuple is the lowest. This tuple cannot be selected again. We loop over the real PE corpus until we obtain a filtered corpus of desirable size (\emph{500k}).

\begin{table}
    \centering
    \begin{tabular}{ | F | F | F | }
   \hline
   Token               &           Count & Percentage \\ 
   \hline
    \texttt{KEEP}      &          326581 &  66.9\% \\
    \texttt{DEL}       &          76725  &  15.7\% \\
    \texttt{"}         &          5170   &  1.1\% \\
    \texttt{,}         &          3249   &  0.7\% \\
    \texttt{die}       &          2461   &  0.5\% \\
    \texttt{der}       &          1912   &  0.4\% \\
    \texttt{zu}        &          1877   &  0.4\% \\
    \texttt{werden}    &          1246   &  0.3\% \\
    \hline
   \hline
   \texttt{KEEP}          &         18367  &  90.4\% \\
   \texttt{DEL}           &           801  &   3.9\% \\
   \texttt{"}             &           199  &   1.0\% \\
   \texttt{>}             &           130  &   0.6\% \\
   \texttt{,}             &            93  &   0.5\% \\
   \texttt{zu}            &            63  &   0.3\% \\
   \texttt{werden}        &            37  &   0.2\% \\
   \texttt{wird}          &            30  &   0.1\% \\
   \hline
\end{tabular}
    \caption{Top 8 edit ops in the target side of the training set for \emph{en-de} (top), and most generated edit ops by our primary (500k + 23k) system on dev set (bottom).}
    \label{table:statsende}
\end{table}

\begin{table*}[t]
\centering
\begin{tabular}{ | c | c | c | c | c | c | c | c | }
    \hline
    \multirow{2}{*}{Model} & \multirow{2}{*}{PE attention} & \multirow{2}{*}{Data} & dev & test 2016 & \multicolumn{2}{c|}{test 2017} & \multirow{2}{*}{Steps} \\
    \cline{4-7}
                           & & & \multicolumn{3}{c|}{TER} & BLEU & \\
    \hline
    \multicolumn{2}{|c|}{Baseline}                 & none            & 24.81 & 24.76 & 24.48 & 62.49 & \\
    \multicolumn{2}{|c|}{SPE}                      & 12k             &       & 24.64 & 24.69 & 62.97 & \\
    \cline{1-7}
    \multicolumn{2}{|c|}{Best 2016 (AMU)}   & 4M + 500k + 12k        & 21.46 & 21.52 & & & \\
    \multicolumn{2}{|c|}{Best 2017 (FBK)}   & 23k + ?                          & & & \textbf{19.60} & 70.07 & \\
    \hline
    \hline
    \multirow{2}{*}{Mono-source} & global & \multirow{3}{*}{12k} & 24.15 & 24.26 & & & 29000 \\
                                 & forced (contr. 1) & & 23.20 & 23.32 & 23.51 & 64.52 & 16600 \\

    \cline{1-1}
    \multirow{2}{*}{Chained} & forced (contr. 2) & & 23.40 & 23.30 & 23.66 & 64.46 & 23600 \\
	\cline{3-3}
                             & forced (primary) & 500k + 12k & 22.77 & \textbf{22.94} & \textbf{23.22} & 65.12 & 119200 \\
    \hline
    \hline
    
	\multirow{2}{*}{Mono-source} & global & \multirow{3}{*}{23k} & 23.60 & 23.55 & & & 47200 \\           
                                 & forced (contr. 1) & & 23.07 & 22.89 & 23.08 & 65.57 & 38800 \\

    \cline{1-1}
    \multirow{2}{*}{Chained} & forced (contr. 2) & & 22.61 & 22.76 & 23.15 & 64.94 & 50400 \\
	\cline{3-3}    
                             & forced (primary) & 500k + 23k & 22.03 & \textbf{22.49} & \textbf{22.81} & 65.91 & 121200 \\
    \hline    
\end{tabular}
    \caption{Results on the \emph{en-de} task. The SPE results are those provided by the organizers of the task (SMT system). The AMU system is the winner of the 2016 APE task \cite{amu}. FBK is the winner of this year's edition. We evaluate our models on \emph{dev} every $200$ training steps, and take the model with the lowest TER. The \emph{steps} column gives the corresponding training time (SGD updates).
    500k + 12k is a concatenation of the 500k synthetic corpus with the 12k corpus oversampled 20 times.
    500k + 23k is a concatenation of 500k with 23k oversampled 10 times.} 
    \label{table:WMT16}

\end{table*}

\subsection{Experimental settings}

We trained mono-source forced models, as well as chained models for both APE tasks. We also trained mono-source models with a global attention mechanism, similar to \cite{cuni} as a measure of comparison to our forced models.

For \emph{en-de}, we trained two sets of models (with the same configuration) on the 12k train set (to compare with 2016 competitors), and on the new (23k) train set.

The encoders are bidirectional LSTMs of size 128. The embeddings have a size of 128.
The first state of the decoder is initialized with the last state of the forward encoder (after a non-linear transformation with dropout). Teacher forcing is used during training (instead of feeding the previous generated output to the decoder, we feed the ground truth).
Like \newcite{bahdanau2015}, there is a maxout layer before the final projection.

We train our models with pure SGD with a batch size of 32, and an initial learning rate of $1.0$.
We decay the learning rate by 0.8 every epoch for the models trained with real PE data, and by 0.5 every half epoch for the models that use additional synthetic data.
The models are evaluated periodically on a \emph{dev} set, and we save checkpoints for the best TER scores.

We manually stop training when TER scores on the dev set stop decreasing, and use the best checkpoint for evaluation on the test set (after about 50k steps for the small training sets, and 120k steps for the larger ones).

Unlike \newcite{amu}, we do not use subword units, as we found them not to be beneficial when predicting edit operations. For the larger datasets, our vocabularies are limited to the 30,000 most frequent symbols.

Our implementation uses TensorFlow \cite{tensorflow}, and runs on a single GPU.\footnote{Our source code, and the configurations used in the experiments are available here: \url{https://github.com/eske/seq2seq/tree/APE}}

\subsection{Results \& Discussion}

\begin{table*}
\centering
\begin{tabular}{ | c | c | c | c | c | c | c | c | }
    \hline
    \multirow{2}{*}{Model} & \multirow{2}{*}{PE attention} & \multirow{2}{*}{Data} & train-dev & dev & \multicolumn{2}{c|}{test 2017} & \multirow{2}{*}{Steps} \\
    \cline{4-7}
                           & & & \multicolumn{3}{c|}{TER} & BLEU & \\

    \hline
    \multicolumn{2}{|c|}{Baseline}                 & none            & 16.11 & 15.58 & 15.55 & 79.54 & \\
    \multicolumn{2}{|c|}{SPE}                      & 24k             &       &       & 15.74 & 79.28 & \\
    \multicolumn{2}{|c|}{Best 2017 (FBK)}          & 24k + ?                               & & & \textbf{15.29} & 79.82 & \\
    \hline
    \multirow{2}{*}{Mono-source} & global & \multirow{3}{*}{24k} & 16.06 & 15.55 &       & & 5200 \\
                                 & forced (contr. 1) &           & 16.05 & 15.57 & 15.62 & 79.48 & 3400 \\

    \cline{1-1}
    \multirow{2}{*}{Chained}     & forced (contr. 2) &           & 16.02 & 15.63 & 15.68 & 79.35 & 7000 \\
    \cline{3-3}
                                 & forced (primary) & 500k + 24k & 15.98 & 15.67 & 15.53 & 79.46 & 27200 \\
    \hline
\end{tabular}
    \caption{Results on the \emph{de-en} task.
    Because the test set was not available before submission, we used a small part (1000 tuples) of the training set as a \emph{train-dev} set. This set was used for selecting the best models, while the provided dev set was used for final evaluation of our models. The \emph{500k + 24k} corpus is a concatenation of our synthetic corpus with the 24k corpus oversampled 10 times.}
    \label{table:WMT17}
\end{table*}

As shown in table~\ref{table:WMT16}, our \emph{forced} (contrastive 1) system gets good results on the \emph{en-de} task, in limited data conditions (\emph{12k} or \emph{23k}). It improves over the MT and SPE baselines, and over the global attention baseline \cite{cuni}.
The chained model, which also uses the source sentence, is able to harness larger volumes of data, to obtain yet better results (primary model). However, it lags behind large word-based models trained on larger amounts of data \cite{amu,amu2017,hokamp}.\footnote{More results are published on the web page of the task:\\ \url{http://statmt.org/wmt17/ape-task.html}}

\begin{figure*}
  \centering
  \subfloat[Global attention 12k train set]{\includegraphics[height=200pt]{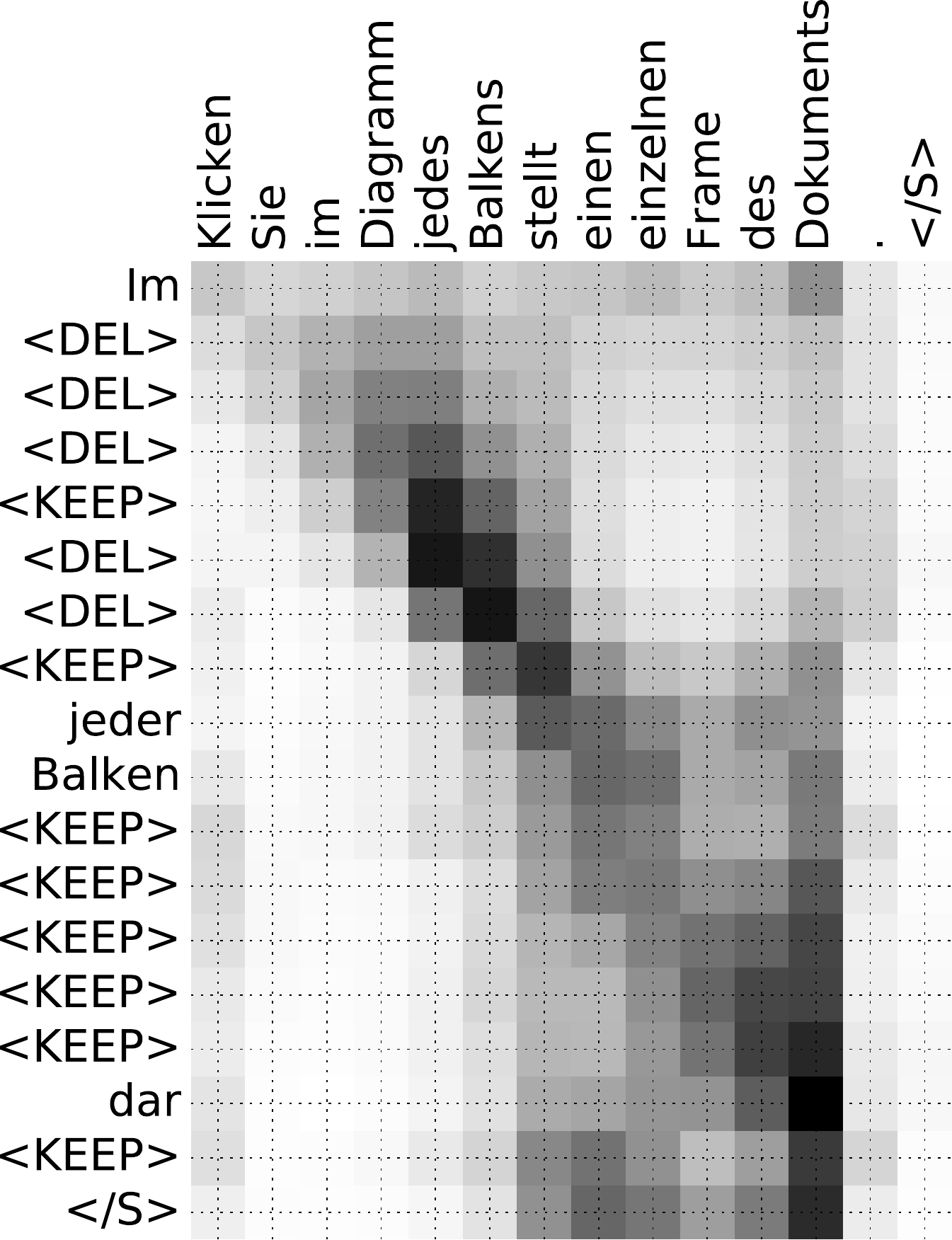}\label{fig:global_12k}}
  \hfill
  \subfloat[Global attention 23k train set]{\includegraphics[height=200pt]{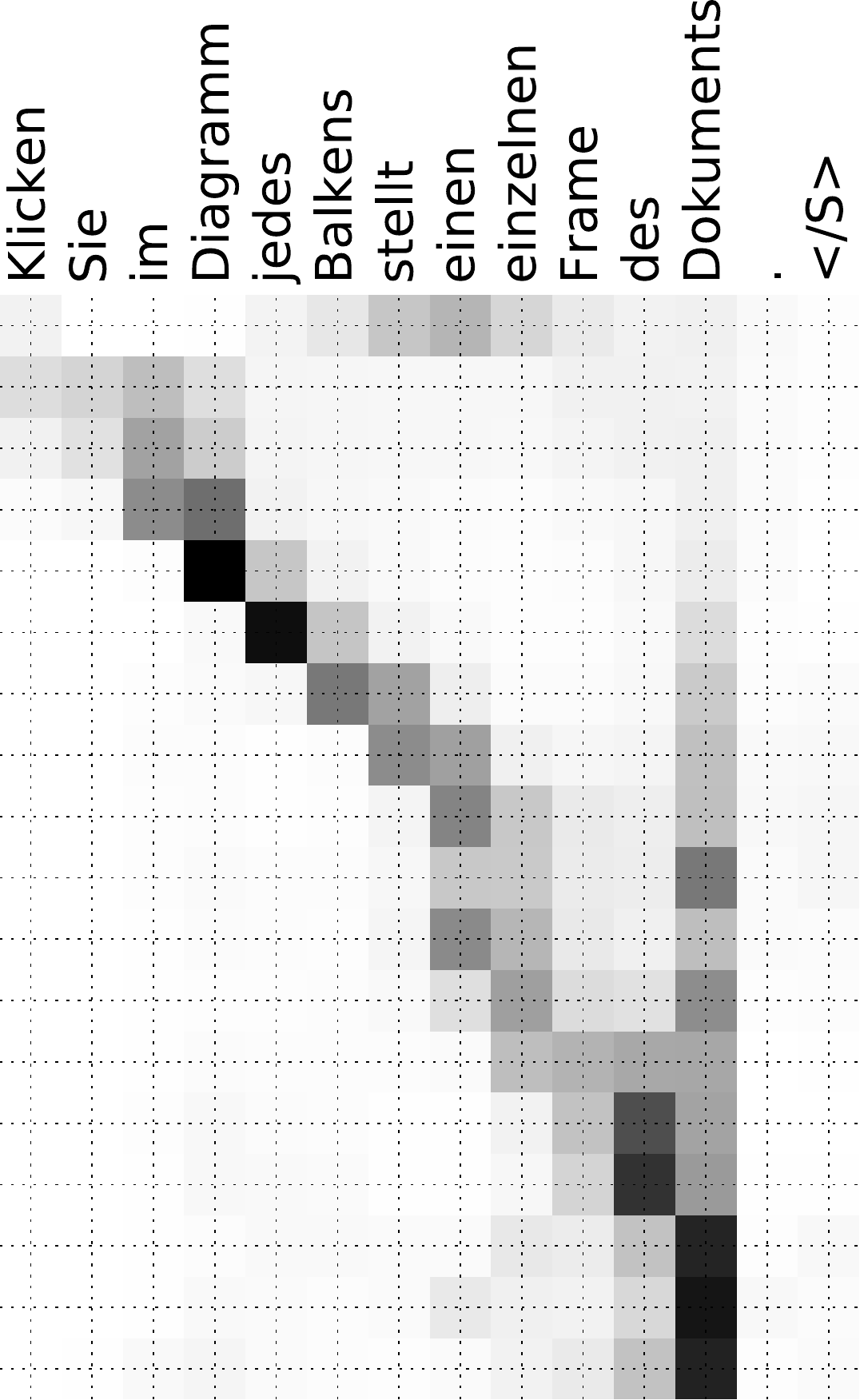}\label{fig:global_23k}}
  \hfill
  \subfloat[Forced attention]{\includegraphics[height=200pt]{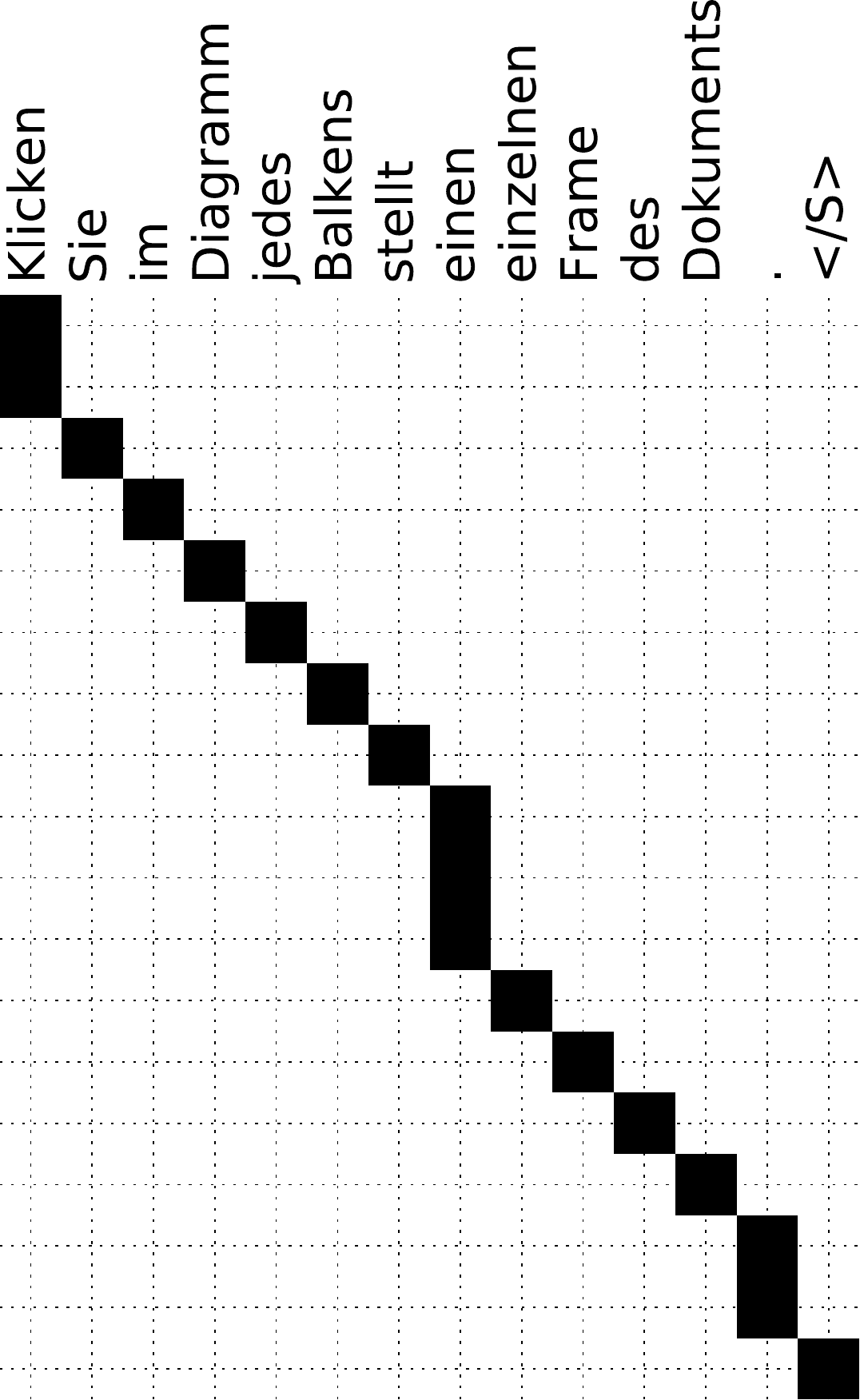}\label{fig:forced}}
  \caption{Alignments of predicted edit operations (\OP{}) with translation hypothesis (\MT{}), on \emph{en-de} dev set, obtained with different attention models.}
  \label{fig:local_vs_global}
\end{figure*}

Figure~\ref{fig:local_vs_global} compares alignments performed by our attention models. We see that the global attention model struggles to learn a meaningful alignment on a small dataset (12k). When more training data is available (23k), it comes closer to our forced alignment.

We see that our good results on \emph{en-de} do not transfer well to \emph{de-en} (see table~\ref{table:WMT17}).
The BLEU scores are already very high (about 16 points above those of the \emph{en-de} data,
and 10 points above the best APE outputs for \emph{en-de}). This is probably due to the translation direction being reversed (because of its rich morphology, German is a much harder target that English). The results obtained with a vanilla SMT system (SPE) seem to confirm this difficulty.

The only reason why our \emph{de-en} systems are able to not deteriorate the baseline, is that they only learned to do nothing, by producing
arbitrarily long sequences of \texttt{KEEP} symbols.
Furthermore, we see that the best results are obtained very early in training, before the models start to overfit and deteriorate the
translation hypotheses on the dev set (see \emph{steps} column).

The difference between our scores on the \emph{de-en} dataset is not statistically significant, therefore we cannot draw conclusions as to which model is the best.
Furthermore, it turns out that our models output almost only \texttt{KEEP} symbols, resulting in sequences almost identical to the \MT{} input, which explains why the scores are so close to those of the baseline (see table~\ref{table:statsdeen}).

Adding substitutions is not particularly useful as it leads to even more data sparsity: it doubles the vocabulary size, and results in less \texttt{DEL} symbols,
and less training feedback for each individual insertion.

\begin{table*}
\centering
\begin{tabular}{ | F | F | F | }
   \hline
   Token                &        Count & Percentage \\ 
   \hline
    \texttt{KEEP}       &       382891 &   78.19\% \\ 
    \texttt{DEL}        &        51977 &   10.61\% \\
    \texttt{the}        &         2249 &    0.46\% \\
    \texttt{,}          &         1691 &    0.35\% \\
    \texttt{of}         &         1620 &    0.33\% \\
    \texttt{to}         &         1022 &    0.21\% \\
    \texttt{a}          &          952 &    0.19\% \\
    \texttt{in}         &          919 &    0.19\% \\
   \hline
\end{tabular}
\hspace{1cm}
\begin{tabular}{ | F | F | F | }
   \hline
   Token                   &     Count & Percentage \\ 
   \hline
    \texttt{KEEP}          &     17861 &   99.62\% \\ 
    \texttt{DEL}           &        52 &    0.29\% \\ 
    \texttt{UNK}           &         4 &    0.02\% \\ 
    \texttt{:}             &         3 &    0.02\% \\ 
    \texttt{the}           &         2 &    0.01\% \\ 
    \texttt{Have}          &         2 &    0.01\% \\ 
    \texttt{A}             &         1 &    0.01\% \\ 
    \texttt{>}             &         1 &    0.01\% \\
   \hline
\end{tabular}
    \caption{Top 8 edit ops in the target side of the training set for \emph{de-en} (left), and most generated edit ops by our primary system on train-dev (right).}
    \label{table:statsdeen}
\end{table*}

\paragraph{Future work}

One major problem when learning to predict edit ops instead of words, is the class imbalance. There are much more \texttt{KEEP} symbols in the training data as any other symbol (see tables~\ref{table:statsende} and~\ref{table:statsdeen}). This results in models that are very good at predicting \texttt{KEEP} tokens (\emph{do-nothing} scenario), but very cautious when producing other symbols.
This also results in bad generalization as most symbols appear only a couple of times in the training data.

We are investigating ways to get a broader training signal when predicting \texttt{KEEP} symbols. This can be achieved either by weight sharing, or by multi-task training \cite{luong2016}.

Another direction that we may investigate, is how we obtain sequences of edit operations (from PE data in another form).
Our edit operations are extracted artificially by taking the shortest edit path between \MT{} and \PE{}.
Yet, this does not necessarily correspond to a plausible sequence of operations done by a human.
One way to obtain more realistic sequences of operations, would be to collect finer-grained data from human post-editors: key strokes, mouse movements and clicks could be used to reconstruct the `true' sequence of edit operations.

Finally, we chose to work at the word level, when a human translator often works at the character level. If a word misses a letter, he won't delete the entire word and write it back.
However, working with characters poses new challenges: longer sequences means longer training time, and more memory usage. Also, it is easier to learn semantics with words (a \emph{character embedding} does less sense). Yet, using characters means more training data, and less sparse data, which could be very useful in a post-editing scenario.

\clearpage
\bibliography{main}
\bibliographystyle{emnlp_natbib}

\end{document}